\documentclass[%
 reprint,
 amsmath,amssymb,
 aps,
]{elsarticle}

\usepackage{graphicx}
\usepackage{dcolumn}
\usepackage{bm}
\usepackage{xcolor}
\usepackage{relsize}
\usepackage{amsmath}
\DeclareMathOperator{\sech}{sech}
\usepackage{amsfonts}
\usepackage{lineno}
\usepackage{lscape}
\usepackage{rotating}
\usepackage{siunitx}
\usepackage{hyperref}

\def\TT{\textcolor{black} }

\begin{document}


\title{Deep learning-enhanced ensemble-based data assimilation for high-dimensional nonlinear dynamical systems}



\author[add1]{Ashesh Chattopadhyay}
\author[add1]{Ebrahim Nabizadeh}
\author[add2,add4]{Eviatar Bach}
\author[add1,add3]{Pedram Hassanzadeh}
\ead{pedram@rice.edu}
\address[add1]{Department of Mechanical Engineering, Rice University, Houston, TX}
\address[add2]{Geosciences Department and Laboratoire de Météorologie Dynamique (CNRS and IPSL), École Normale Supérieure and PSL University, Paris, France}
\address[add4]{Division of Geological and Planetary Sciences, California Institute of Technology, Pasadena, CA}

\address[add3]{Department of Earth, Environmental and Planetary Sciences, Rice University, Houston, TX}






\begin{abstract}
Data assimilation (DA) is a key component \TT{of many forecasting models in} science and engineering. DA allows one to estimate \TT{better} initial conditions \TT{using} an imperfect dynamical model of the system and noisy/sparse observations \TT{available} from the system. Ensemble Kalman filter (EnKF) is a DA algorithm that is widely used in applications involving high-dimensional nonlinear dynamical systems. However, EnKF requires evolving large ensembles of \TT{forecasts} using the dynamical model of the system. This often becomes computationally intractable, especially when the number of states of the system is very large, e.g., \TT{for weather prediction}. \TT{With small ensembles,} the estimated background error covariance matrix in the EnKF algorithm suffers from sampling error, \TT{leading} to an erroneous estimate of the analysis state (initial condition for the next forecast cycle).  In this work, we propose hybrid ensemble Kalman filter (H-EnKF), \TT{which is applied to a two-layer quasi-geostrophic flow system as a test case}. \TT{This framework} utilizes a pre-trained deep learning-based data-driven surrogate \TT{that} inexpensively generates and evolves a large data-driven ensemble of the states of the \TT{system} to accurately compute the background error covariance matrix \TT{with less} sampling error. \TT{The H-EnKF framework estimates a better initial condition without the need for any ad-hoc localization strategies. H-EnKF can be extended to any ensemble-based DA algorithm, e.g., particle filters, which are currently difficult to use for high-dimensional systems.}  
\end{abstract}

\maketitle


\section{Introduction}

    Data assimilation (DA) is an indispensable component \TT{in many of the forecasting models used for applications in science and engineering} \cite{kalnay2003atmospheric,law2015data,asch2016data,morzfeld2018feature,morzfeld2019gaussian}. \TT{DA allows one to estimate better initial conditions for a system from which noisy and sparse observations are available, along with an imperfect dynamical model of the system.} These initial conditions are \TT{then} used by the dynamical model to predict the future states of the system. \TT{The} quality of future forecasts \TT{strongly} depends on the accuracy
    of the initial conditions. This is especially important in chaotic systems, where even a small error in an initial condition can result in drastically different forecasts \TT{\citep{lorenz1996predictability}}. In such systems, an accurate initial condition is of the utmost importance for predictive dynamical models to provide accurate forecasts. \TT{DA is critical} in various areas of engineering and sciences, such as \TT{weather prediction \cite{carrassi2018data,chen2022efficient,gleiter2022ensemble}}, environmental and geophysical flows  \cite{belyaev2018optimal,d2012hpc,arcucci2019optimal}, combustion systems \cite{yi2008online,bell2019bayesian,croci2021data}, aeronautics \cite{brunton2021data}, hydrology \citep{liu2007uncertainty}, acoustics \cite{maday2015parameterized}, and fluid mechanics \cite{gu2007iterative, habibi2021integrating}.

     There are two main categories of DA algorithms: ensemble-based methods and variational methods. Ensemble-based DA algorithms such as ensemble Kalman filter (EnKF) were proposed for estimating \TT{better} initial conditions from noisy observations \TT{assuming} Gaussian observation noise \cite{evensen1994sequential}. Several other DA algorithms such as variational methods (e.g., 3D-Var and 4D-Var) or a combination of ensemble and variational algorithms \TT{also exist} \cite{fletcher2017data,bannister2017review}. 4D-Var requires obtaining the adjoint of the \TT{system's} dynamical model, which is often a difficult and non-trivial task. Moreover, 4D-Var optimizes a cost function which can be computationally expensive. Ensemble-based algorithms do not require obtaining adjoints \TT{but require evolving a large ensemble of forecasts of the dynamical system}. In this paper, we will focus on ensemble-based algorithms for DA.

    A major challenge in EnKF is generating and evolving a large number of ensemble members of the dynamical model in time. The accuracy of the background error covariance matrix in EnKF (which affects the performance of EnKF) \TT{depends} on the ensemble size \cite{kondo2016impact}. For an accurate estimation of the background error covariance matrix, the number of ensemble members should be of the order of the number of states, \TT{$s$,} in the system, \TT{which can be large} (e.g., \TT{in the weather system, $s$} $\approx O(10^7)-O(10^8)$).  However, evolving such large ensembles over multiple time steps becomes computationally intractable. Thus, fewer ensembles are typically used in practice ($\approx O(50)$ in operational weather models \cite{leutbecher2019ensemble}). These covariance matrices, generated from a smaller number of ensemble members, are rank-deficient and suffer from sampling error that degrades the quality of the estimated initial condition (often referred to as the ``analysis state"). For this reason, various ad-hoc localization strategies \TT{have been proposed} to remove spurious long-range spatial correlations in the covariance matrix \cite{asch2016data}. However, in this process, one may also remove physically-consistent long-range spatial correlations and this can adversely affect the performance of EnKF \TT{and thus, the quality of forecasts}~\cite{miyoshi201410}. There are other methods to estimate the background error covariance matrix without evolving a large ensemble, e.g., the stochastic Galerkin method \citep{janjic2021test}. 
    
     In recent years, we have seen an increase in interest at the intersection of  machine learning (ML) and DA for applications in dynamical systems. \TT{For example,} in Yang \textit{et al.} \citep{yang2021machine}, a generative model was used to facilitate ensemble generation for analog-based DA. In our previous work \citep{chattopadhyay2021towards}, we have shown that a data-driven weather forecasting model can be integrated with a sigma-point ensemble Kalman filter to obtain accurate initial conditions for forecasting. Tsuyuki \textit{et al.}~\citep{tsuyuki2022nonlinear} integrated ML with an EnKF to perform state estimation in a nonlinear dynamical system using a small number of ensemble members. Maulik \textit{et al.} \citep{maulik2022aieada} used 4D-Var-based DA with ML for forecasting in high-dimensional dynamical systems. Penny \textit{et al.} \citep{penny2022integrating} used a recurrent neural network and 4D-Var for scalable state estimation. Chen \textit{et al.} \citep{chen2021bamcafe} showed the application of DA in ML-based forecasts \TT{in complex turbulent flows with partial observations}. \TT{There are several other studies that have shown different applications, where ML has been integrated with DA, e.g., predicting subgrid-scale processes in multi-scale chaotic systems~\citep{brajard2020combiningSGS,pawar2020long}, closed-form equation discovery of model error \citep{mojgani2021closed}, etc.}

    In this paper, we propose hybrid ensemble Kalman filter (H-EnKF), a hybrid algorithm for enhancing the performance of EnKF without resorting to localization strategies. \TT{H-EnKF} leverages deep learning to build a data-driven surrogate of the dynamical model of the system. This surrogate is used to generate and evolve a large \TT{number of} data-driven ensemble members, \TT{$O(s)$}, to compute the background error covariance matrix of the dynamical system. The data-driven model is trained on the full state of the system and serves as a computationally cheap surrogate to predict the evolution of the states, \TT{and is used just for estimating the background error covariance matrix.} While the less-accurate but large data-driven ensembles are used to compute the background error covariance matrix with low sampling error, the numerical model of the system is used to generate and evolve a small number of ensemble members, $g$, where $g \ll s$ (note that $g$ could be just $1$). \TT{This small number of ensemble members} are used to compute an accurate background forecast state to be used in the H-EnKF algorithm. \TT{In order to demonstrate the performance of H-EnKF, we have applied it to a well-known test-bed for fully turbulent geophysical flows: the two-layer quasi-geostrophic (QG) system \citep{lutsko2015applying,nabizadeh2019size}.}
    
    
    The remainder of the paper is organized \TT{into several sections.}~\TT{In section~\ref{sec:methods}, we describe the QG system, the numerical solver used to simulate the system, the data-driven surrogate model, and the EnKF algorithm. In section~\ref{sec:HENKF}, we introduce our proposed H-EnKF algorithm. Section~\ref{sec:metrics} describes the metrics to evaluate the performance of EnKF and H-EnKF. In section~\ref{sec:results}, the performance of the algorithms, in terms of accuracy and cost, is presented, followed by summary and discussion in section~\ref{sec:discussion}.}
    
\section{Method}
\label{sec:methods}
\subsection{Two-layer QG System}
The dimensionless dynamical equations of the \TT{two}-layer QG flow have been developed following Lutsko \textit{et al.}~\cite{lutsko2015applying} and Nabizadeh \textit{et al}.~\cite{nabizadeh2019size}. The system consists of two constant density layers with a $\beta$-plane approximation in which the meridional temperature gradient is relaxed towards an equilibrium profile. The system's equation is

\begin{align}
\label{eq_QG}
\begin{split}
    \frac{\partial q_k}{\partial t}+J(\psi_k,q_k)=-\frac{1}{\tau_d}(-1)^{k}\left(\psi_1-\psi_2-\psi_R\right) \\
    -\frac{1}{\tau_f}\delta_{k2}\nabla^2\psi_k - \nu \nabla^{8}q_k.
       \end{split}
\end{align}
Here, $q$ is potential vorticity
\begin{align}
\label{q_def}
q_k=\nabla^2\psi_k +(-1)^{k}\left(\psi_1-\psi_2\right)+\beta y,
\end{align}
where $\psi_k$ is the streamfunction of the system.
In Eqs.~(\ref{eq_QG}) and~(\ref{q_def}), $k$ denotes the upper ($k=1$) and lower ($k=2$) layers.~$\beta$ is the $y-$gradient of the Coriolis parameter.~$\tau_d$ is the Newtonian relaxation time scale and $\tau_f$ is the Rayleigh friction time scale, which only acts on the lower layer. $\delta_{k2}$ is the Kronecker $\delta-$ function.~$J$ denotes the Jacobian.~$\nu$ denotes the hyperdiffusion coefficient. We have introduced a baroclinically unstable jet at the center of a \TT{zonally} periodic channel by setting $\psi_1-\psi_2$ to be equal to a hyperbolic secant centered at $y=0$. When eddy fluxes are absent, $\psi_2$ is identically zero, making zonal velocity in the upper layer, $u_1(y)=-\frac{\partial \psi_1}{\partial y}=-\frac{\partial \psi_R}{\partial y}$, where we set
\begin{align}
\label{init_eq}
-\frac{\partial \psi_R}{\partial y}=\sech^2\left(\frac{y}{\sigma}\right).
\end{align}
$\sigma$ \TT{is} the width of the jet. Parameters of the model are set following previous works~\cite{lutsko2015applying,nabizadeh2019size};~$\beta= 0.19$, $\sigma=3.5$, $\tau_f=15$, and $\tau_d=100$. 

To non-dimensionalize the equations, we have used the maximum strength of the equilibrium velocity profile as the velocity scale ($U$) and the deformation radius ($L$) for the length scale. The system’s time scale ($L/U$) is referred to as the “advective time scale” ($\tau_{adv}$) which is approximately $6$h in this system.

\subsection{Numerical Solver}
\label{sec:numerical}
The spatial discretization is spectral in both $x$ and $y$, where we have retained $96$ and $192$ Fourier modes, respectively. The length and width of the domain are equal to $46$ and $68$, respectively. \TT{Sponge layers are applied to the northern and southern boundaries.} Note that the domain is wide enough \TT{for the sponges to} not affect the dynamics. Here $5 \tau_{adv} \approx$ $1$ Earth day $\approx 200 \Delta t$, where $\Delta t=0.025$ is the time step of the leapfrog time integrator used in the numerical scheme.

\subsection{The Data-driven Model: \TT{U-NET}}
\label{sec:DD}
\begin{figure*}[ht]
\includegraphics[width = \textwidth]{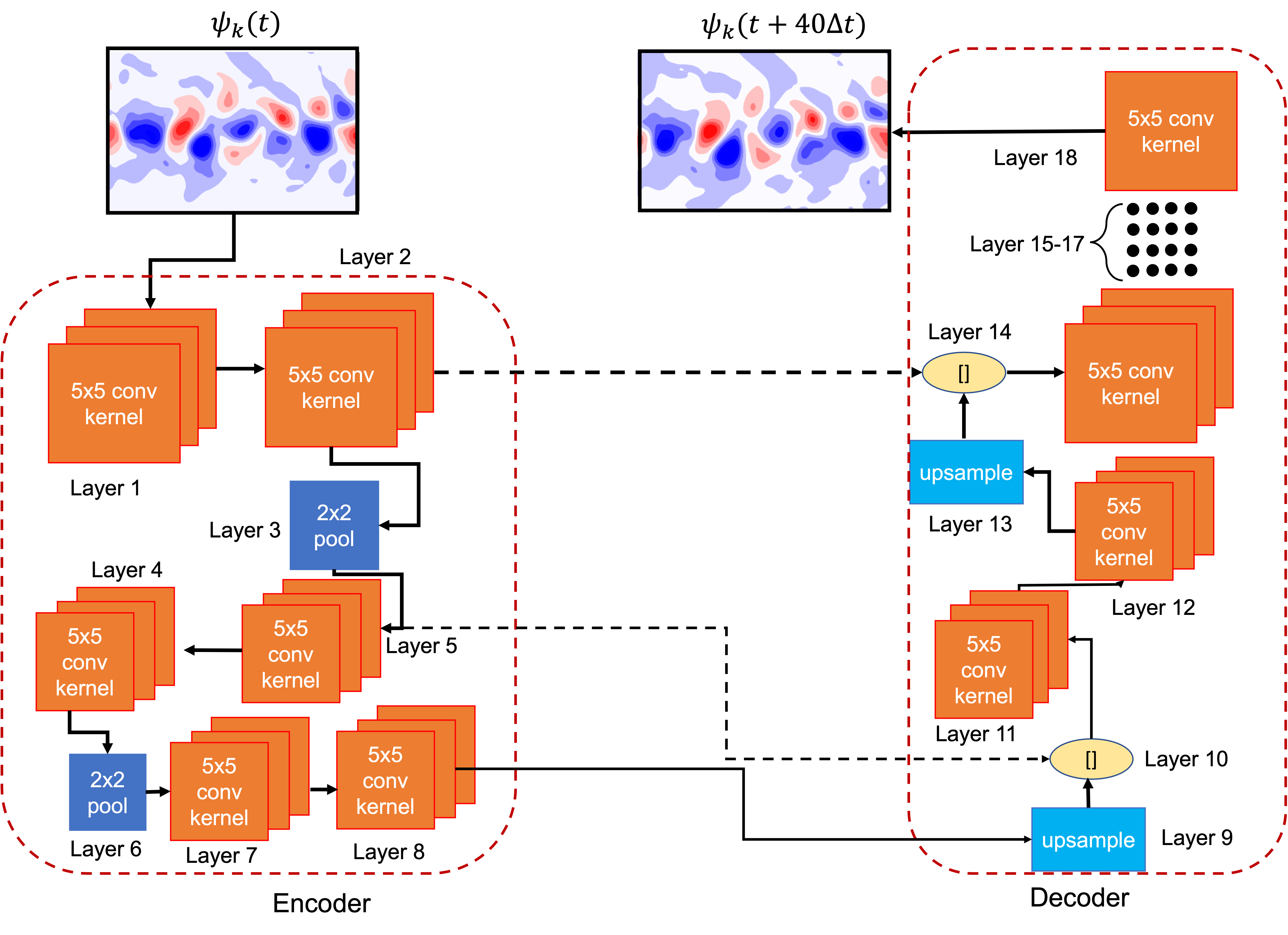}
\caption{\label{fig:U-STN} \TT{Schematic of the U-NET model used as a surrogate for data-driven prediction of $\psi_k (t)$ in the two-layer QG system. The input to the model is the system's full state, $\psi_k(t)$ and the output is $\psi_k(t+40\Delta t)$, where $k=\{1,2\}$.} The detailed information of the number of layers, their index number, number of filters, and size of convolutional kernels after hyperparameter optimization is given in Table~\ref{tab:UNET}. Here, $40 \Delta t \approx 6h$.}
\end{figure*}

\TT{To build a data-driven surrogate model to approximate the QG dynamics, we adopt a U-NET architecture \citep{ronneberger2015u}. The choice of this architecture is inspired by our previous work in data-driven weather forecasting \citep{chattopadhyay2021towards}.} The details of the architecture are provided in Table~\ref{tab:UNET} and a schematic is \TT{shown} in Fig.~\ref{fig:U-STN}. The U-NET allows one to extract small-scale features in the encoder that are directly passed into the decoder as skipped connections. This \TT{ has been shown to result in improved} performance on turbulent flow prediction \cite{wang2020towards}. The U-NET architecture is trained on $N_{tr}$ \TT{noise-free samples} (we have performed experiments with $N_{tr}=10^5$ and $N_{tr}=10^4$ samples; see section~\ref{sec:DD_perf}) of the system's state, i.e., streamfunction, $\psi_k(t)$, $k=\{1,2\}$. \TT{The input} to the network is $\psi_k(t)$ and the target is $\psi_k(t+\gamma\Delta t)$. \TT{We have shown in a previous study~\citep{chattopadhyay2021towards} that large values of $\gamma$ improve the prediction horizon of the data-driven model over small values of $\gamma$. Hence, $\gamma > 1$ would result in a longer prediction horizon than $\gamma = 1$. However, in the H-EnKF algorithm (see section~\ref{sec:HENKF} for more details) an accurate computation of the background error covariance matrix with data-driven ensembles requires one to evolve the data-driven model for a sufficient number of time steps as well. Since we obtain observations from the system at every $200 \Delta t$ (1 DA cycle),  $\gamma$ has to be smaller than $200$. Hence, the aim is to choose a $\gamma$ such that the U-NET has a good prediction horizon while the data-driven ensembles can evolve sufficiently to provide an accurate covariance matrix as well. In this paper, we have chosen $\gamma=40$, such that $\gamma \Delta t$ matches $\tau_{adv}$ ($\approx 6$h) of the QG system.}

The samples of $\psi_k(t)$ are obtained from numerically solving Eq.~(\ref{eq_QG}) and Eq.~(\ref{q_def}). Each sample corresponds to a different time ($t$). The training and testing sets are obtained from independent simulations starting from different random initial conditions so that there is no correlation between the training and testing sets. The hyperparameters of the U-NET \TT{are} determined after extensive trial and error. 

As shown by us and others in previous studies, the performance of the U-NET can be improved by incorporating the symmetries in the system inside the architecture \cite{chattopadhyay2020deep,chattopadhyay2021towards} or by using physics-based regularizers in the loss function of the U-NET \cite{wang2020incorporating}. \TT{However, in the H-EnKF framework, we only need the U-NET to predict short-term evolution ($200 \Delta t \approx 1$ Earth day into the future, more details in section~\ref{sec:HENKF}). For such a short-term forecast, the performance of the U-NET remains roughly the same as compared to an architecture with physics-constrained loss functions or imposed symmetries.} For more complicated problems (such as weather \TT{forecasting}) that might benefit from enforcing physics within the data-driven architecture, see the comprehensive review by Kashinath \textit{et al.} \citep{kashinath2021physics}. \TT{Note that other types of neural architectures can also be used to build the surrogate model, e.g., neural operator-based models, which have recently been shown to achieve state-of-the-art performance in weather forecasting \citep{jpathak2022}.}

\begin{table}[t]
    \centering
    \begin{tabular}{|c|c|c|c| p{0.1\textwidth}|}
    \hline
          \textbf{Number}& \textbf{Layer} & \textbf{Number of Filters} \\
         \hline
         1 & $5\times 5$ 2D Convolution & {32} \\
         \hline
         2 & $5\times 5$ 2D Convolution & 32  \\
         \hline
         3 & $2\times 2$ Max Pooling & --\\
         \hline
         4 & $5\times 5$ 2D Convolution & 32 \\
         \hline
         5 & $5\times 5$ 2D Convolution & 32\\
         \hline
         6 & $2\times 2$ Max Pooling & -- \\
         \hline
         7 & $5\times 5$ 2D Convolution & 32 \\
         \hline
         8 & $5\times 5$ 2D Convolution & 32\\
         \hline
         9 & Up-sampling & --\\
         \hline
         10 & Concatenation & --\\
          \hline
         11 & $5\times 5$ 2D Convolution & 32 \\
          \hline
         12 & $5\times 5$ 2D Convolution & 32\\
          \hline
         13 & Up-sampling & --\\
          \hline
         14 & $5\times 5$ 2D Convolution & 32\\
          \hline
         15 & Concatenation & --\\
          \hline
         16 & $5\times 5$ 2D Convolution & 32\\
          \hline
         17 & $5\times 5$ 2D Convolution & 32 \\
          \hline
         18 & $5\times 5$ 2D Convolution & 2\\
          \hline
    \end{tabular}
    \caption{Number of layers and filters in the U-NET architecture used as the data-driven surrogate model.}
    \label{tab:UNET}
\end{table}

\subsection {Data Assimilation with a Stochastic EnKF}
\label{sec:EnKF}
\TT{In this section, we describe DA with stochastic EnKF. Stochastic EnKF requires a dynamical model of the system, also called ``background forecast model", represented by $M$. We further assume that an ensemble of noisy observations, $\psi^j_{obs}$, where $j$ is the index of the ensemble member, is obtained by adding Gaussian white noise to the observations. Here, we assume that the observation noise distribution can be represented as a standard Gaussian with zero mean and standard deviation of $\sigma_{obs}$. In this paper, we have considered $\sigma_{obs}=0.1$ ($10\%$ of standard deviation of $\psi_k(t)$, $1.0$). Throughout the rest of the paper, we will drop the word ``stochastic", assuming that an ensemble of noisy observations is available as opposed to a single noisy observation.} $M$ evolves an ensemble of state vectors $\psi^j$ from $t$  to $t+\Delta t$ starting from a noisy initial condition, $\psi(t_0)$.  Let us assume that $j \in \{1,2,\cdots n \}$, where $n$ denotes the number of ensemble members. Here, and throughout the rest of the paper, we have dropped the subscript $k$ (the index for the two layers in the system) for clarity unless it is necessary (e.g., in Eq.~(\ref{eq:RMSE}) and Eq.~(\ref{eq:ACC})) keeping in mind that all computations take place on both layers. Furthermore, we assume that assimilation of observations takes place at every $\alpha \Delta t$ (in this paper, we choose $\alpha = 200$, i.e., DA occurs every day). 

The ensemble of state vectors is generated by adding Gaussian white noise with zero mean and standard deviation of $\sigma_b$  to the noisy state vector at time $t_0$, $\psi(t_0)$:

\begin{eqnarray}
\label{eq:gen_ens}
\psi^j(t_0) = \psi(t_0)+\mathcal{N}\left(0,\sigma_b^2\right).
\end{eqnarray}
The value of $\sigma_b$ has been chosen after significant trial and error for both EnKF and H-EnKF algorithms separately to obtain the best performance (see section~\ref{fig:H-ENKF_perf}). The evolution of the ensemble of state vectors, $\psi^j(t)$, \TT{at any time $t$,} can be written in discrete form:

\begin{eqnarray}
\label{eq:forward}
\psi^j(t+\alpha\Delta t)=\underbrace{M \circ M \circ \cdots  M}_{\alpha  }\left[\psi^j(t)\right]. 
\end{eqnarray}
Then, we compute the background error covariance matrix using the ensembles that are obtained at $\alpha \Delta t$:

\begin{equation}
\label{eq:P_b}
\begin{aligned}
  \mathbf{P} = \mathbb{E} \left[ \left(\psi^j(t+\alpha\Delta t)-
   \overline{\psi}(t+\alpha \Delta t)\right) \right. \\
   \left. \left(\psi^j(t+\alpha\Delta t)-
   \overline{\psi}(t+\alpha \Delta t)\right)^T \right],
   \end{aligned}
\end{equation}
where $\overline{\psi}(t+\alpha\Delta t)$ is the mean over the $n$ ensemble members and $\mathbb{E}$ is the sample expectation operator. The Kalman gain is computed using $\mathbf{P}$ as:

\begin{eqnarray}
\label{eq:kalman_gain}
\mathbf{K}=\mathbf{P}\left(\mathbf{P}+\sigma_{obs}\mathbf{I}\right)^{-1}.
\end{eqnarray}
\TT{Here, the observation operator, $H$, is $\mathbf{I}$.} However, we can extend Eq.~(\ref{eq:kalman_gain}) to nonlinear $H$ as well. Finally, using the ensemble of noisy observations, $\psi_{obs}^j(t+\alpha \Delta t)$, the noise-reduced analysis state is computed as:

\begin{eqnarray}
\label{eq:analysis}
\begin{split}
\psi_{a}^j(t+\alpha\Delta t) = \psi^j(t+\alpha\Delta t)-\\
\mathbf{K}\left( \psi_{obs}^j(t+\alpha\Delta t)-\psi^j(t+\alpha\Delta t) \right),
\end{split}
\end{eqnarray}
where $\psi_a^j(t+\alpha\Delta t)$ is the ensemble of analysis states.\TT{~$\overline{\psi_a}(t+\alpha\Delta t)$ is a noise-reduced estimate of the state of the system. Hence, $\overline{\psi_a}(t+\alpha\Delta t)$ is used as the initial condition by the dynamical model, $M$, for free forecasting. Moreover, the ensemble of analysis states, $\psi_a^j(t+\alpha\Delta t)$, can be used for probabilistic forecasting as well, see e.g., Holstein \textit{et al.} \citep{stael1971experiment}.}

\section {Proposed DA Algorithm: H-EnKF}
\label{sec:HENKF}
\begin{sidewaysfigure}
\includegraphics[width=54pc,angle=0,trim={0cm 0cm 0cm 0cm},clip]{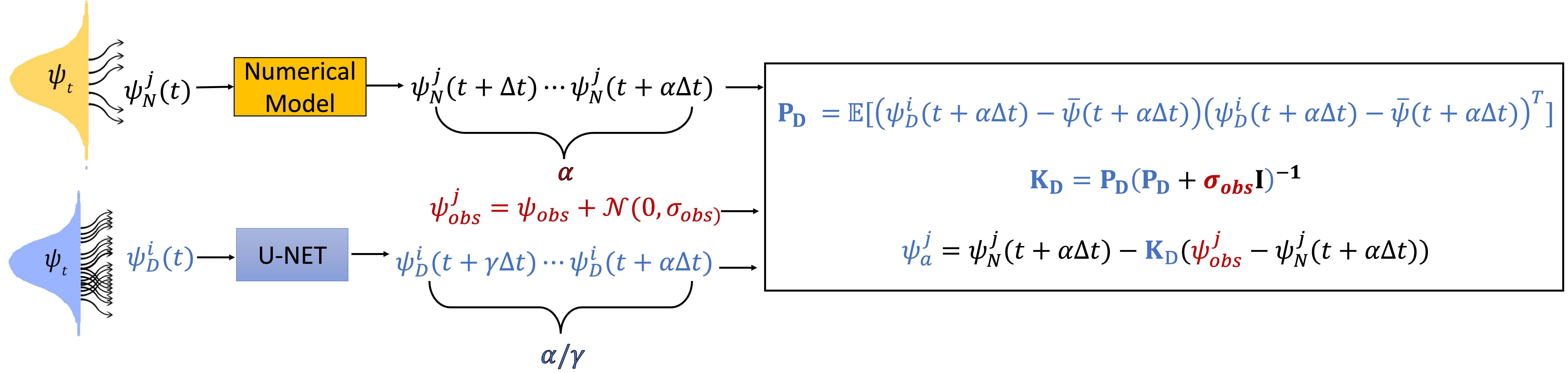}
\caption{\label{fig:H-ENKF} Schematic of the H-EnKF framework. Noisy initial condition $\psi(t_0)$ is perturbed with $\mathcal{N}\left(0, \sigma_b^2 \right)$ to generate two sets of ensembles: $1)$ $\psi_D^i(t)$ with $n_D$ ensembles where $n_D \approx O(1000)$ and $2)$ $\psi_N^j(t)$ with $n_N$ ensembles where $n_N \approx O(10)$. A pre-trained U-NET predicts the evolution of each of the $n_D$ ensembles autoregressively for $\alpha\Delta t$. Similarly, the numerical solver for the QG system evolves the $n_N$ ensembles of $\psi_N^j(t)$ for $\alpha\Delta t$. At this point, a noisy observation of $\psi_{obs}$ is perturbed with $\mathcal{N}\left(0, \sigma_{obs}^2 \right)$ to generate $n_N$ ensembles. Here, an EnKF algorithm computes the background covariance matrix $\mathbf{P_D}$ using the ensembles evolved with the U-NET and finally produces the analysis state ensembles $\psi^j_a$ using the background forecast $\psi_N^j(t)$ from the numerical model. While the analysis ensembles are carried forward and evolved by the numerical model, the U-NET needs the ensembles to be \textit{restarted} by perturbing the ensemble-averaged analysis state using Gaussian noise (zero mean and  $\sigma_b$ standard deviation) generating $n_D$ ensembles to be evolved over the next $\alpha \Delta t$. In this paper, we have chosen $\alpha = 200$ and $\gamma=40$. Here, $40 \Delta t \approx 6h$ and $200 \Delta t \approx 1$ Earth day.}
\end{sidewaysfigure}

The major challenge with using EnKF is computing the background error covariance matrix, $\mathbf{P}$, using Eq.~(\ref{eq:P_b}). An accurate computation of $\mathbf{P}$ requires $n$ to be large, typically the same order as that of the dimension of $\psi(t)$. However, this makes the evaluation of Eq.~(\ref{eq:forward}) computationally expensive. This is especially true for high-dimensional systems where $M$ is an expensive numerical model. Hence, traditional applications of EnKF uses small values of $n$, which induces sampling error in $\mathbf{P}$, resulting in spurious long-range correlations. \TT{In this section, we show how the U-NET-based surrogate model is used in developing H-EnKF. A schematic of H-EnKF is shown in Fig.~\ref{fig:H-ENKF}}. 


We denote the U-NET-based surrogate model as $M_D$, which evolves the state $\psi(t)$ to $\psi(t+\gamma\Delta t)$ (in this paper, $\gamma=40$). Since $M_D$ is already trained, it is computationally inexpensive during inference. Hence, one can afford to evolve a very large number of ensemble members, $O(1000)$, with $M_D$. We denote the number of ensemble members that are evolved with $M_D$ as $n_D$. Similarly, we denote the number of ensemble members evolved with the numerical model ($M_N$) as $n_N$. Note that for all practical problems involving high-dimensional systems, one can computationally afford a small number of $n_N$ but a very large number of $n_D$ ($n_N \ll  n_D$). Similar to section~\ref{sec:EnKF}, DA occurs at every $\alpha \Delta t$ and we assume that $\alpha$ is an integer multiple of $\gamma$.

In H-EnKF, we first evolve a large data-driven ensemble using $M_D$:
\begin{eqnarray}
\label{eq:forward_DD}
\psi_D^i(t+\alpha\Delta t)=\underbrace{M_D \circ M_D \circ \cdots  M_D}_{\alpha/\gamma}\left[\psi_D^i(t)\right],
\end{eqnarray}
where $i \in \{1,2,3,\cdots n_D\}$. At the same time, we evolve a small number of numerical ensemble members using $M_N$:
\begin{eqnarray}
\label{eq:forward_NM}
\psi_N^j(t+\alpha\Delta t)=\underbrace{M_N \circ M_N \circ \cdots  M_N}_{\alpha}\left[\psi_N^j(t)\right],
\end{eqnarray}
where $j \in \{1,2,3,\cdots n_N\}$. In both Eq.~(\ref{eq:forward_DD}) and Eq.~(\ref{eq:forward_NM}), the ensembles are generated with Gaussian white noise as shown in Eq.~(\ref{eq:gen_ens}), . At this point, we compute the background error covariance matrix, $\mathbf{P_D}$, using the $n_D$ ensemble members evolved by $M_D$:

\begin{equation}
\label{eq:P_b_HENKF}
\begin{aligned}
  \mathbf{P_D} = \mathbb{E} \left[ \left(\psi_D^i(t+\alpha\Delta t)-
   \overline{\psi}_D(t+\alpha \Delta t)\right) \right. \\
   \left. \left(\psi_D^i(t+\alpha\Delta t)-
   \overline{\psi}_D(t+\alpha \Delta t)\right)^T \right].
   \end{aligned}
\end{equation}
Similar to Eq.~(\ref{eq:P_b}), $\overline{\psi}_D(t+\alpha \Delta t)$ denotes the mean of $\psi_D^i(t+\alpha\Delta t)$ over $n_D$ ensemble members. Since $n_N \ll n_D$, $\mathbf{P_D}$ calculated with $n_D$ ensemble members would have much lower sampling error as compared to the one computed with $n_N$ ensemble members. Then, we compute Kalman gain, $\mathbf{K_D}$, as:
\begin{eqnarray}
\label{eq:kalman_gainD}
\mathbf{K_D}=\mathbf{P_D}\left(\mathbf{P_D}+\sigma_{obs}\mathbf{I}\right)^{-1}.
\end{eqnarray}
Then we compute the ensemble of analysis states as:

\begin{eqnarray}
\label{eq:analysis_HENKF}
\begin{split}
\psi_{a}^j(t+\alpha\Delta t) = \psi_N^j(t+\alpha\Delta t)-\\
\mathbf{K_D}\left( \psi_{obs}^j(t+\alpha\Delta t)-\psi_N^j(t+\alpha\Delta t) \right),
\end{split}
\end{eqnarray}
where $\psi_a^j(t+\alpha\Delta t)$ is the ensemble of analysis states. Note that in Eq.~(\ref{eq:analysis_HENKF}), we have used $n_N$ ensemble members of the background forecast state which are more accurate (since $M_N$ is a more accurate numerical model that integrates physical equations as compared to the data-driven model, $M_D$). However, $\mathbf{P_D}$ is obtained from the large number ($n_D$) of ensemble members from $M_D$ to alleviate issues related to sampling error and spurious long-range correlations. 

\TT{In H-EnKF, the $n_N$ ensemble members (obtained as the analysis states) keep evolving for future DA cycles with the numerical model, as is done in a standard EnKF. However, after every DA cycle, we reinitialize the $n_D$ ensemble members using Eq.~(\ref{eq:gen_ens}), where $\psi(t_0)$ is replaced with $\overline{\psi_a}(t+\alpha\Delta t)$ for $t>t_0$. This is because the data-driven models' forecasts do not remain stable for long time scales~\citep{chattopadhyay2020deep,keisler2022forecasting,chattopadhyay2022long}.  } 

\TT{It is important to note that, one can also use the data-driven model's prediction to compute the background forecast state, i.e., a fully data-driven model without the need for any numerical integration~\citep{chattopadhyay2021towards}. However, the quality of such forecasts would depend on how frequently we perform DA. If the frequency of DA is small, i.e., we evolve the data-driven model for a long period of time before performing DA, the quality of the data-driven forecasted state would degrade. That would lead to inaccuracies in the computation of the analysis state.}

\section{Metrics for measuring performance}
\label{sec:metrics}

We define two metrics for measuring the performance of a model: relative error ($E_k(t)$) and Anomaly Correlation Coefficient ($\text{ACC}_k$) for each layer, $k$. Details on computing these metrics are given in~\ref{secA:error}.

\section{Results}
\label{sec:results}
\subsection{Performance of U-NET for Fully Data-driven Prediction}
\label{sec:DD_perf}
\begin{sidewaysfigure}
\includegraphics[width = \textwidth,angle=0,trim={0cm 2.0cm 0cm 0cm},clip]{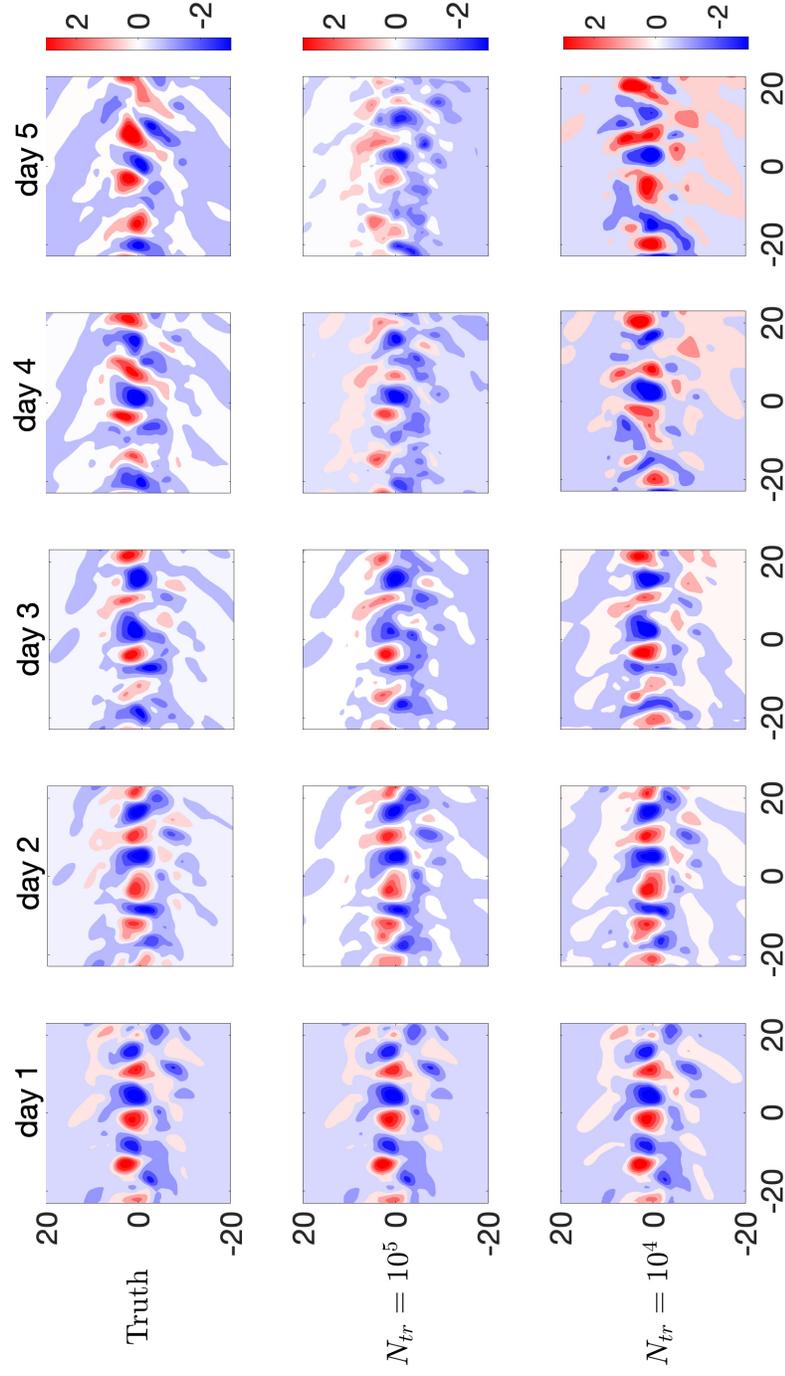}
\caption{\label{fig:snapshot_DD} Predicted patterns of time-mean removed $\psi_1(t)$ anomalies by the U-NET trained on $N_{tr}=10^5$ and $N_{tr}=10^4$  training samples as compared to the truth (obtained from numerical simulation). Note that only part of the latitudinal extent of the domain is shown.}
\end{sidewaysfigure}

First, we show the performance of fully data-driven predictions with U-NET trained on $N_{tr}$ samples of $\psi_k(t)$. Figure~\ref{fig:snapshot_DD} shows the predicted patterns of the time-mean removed anomalies of $\psi_1(t)$ with U-NET trained on $N_{tr}=10^4$ and $N_{tr}=10^5$ samples. Qualitatively good performance up to day $3$ can be seen in Fig.~\ref{fig:snapshot_DD} with both $N_{tr}=10^4$ and $N_{tr}=10^5$ samples.

\begin{sidewaysfigure}

\includegraphics[width = \textwidth,angle=0,trim={2cm 2.0cm 3cm 3cm},clip]{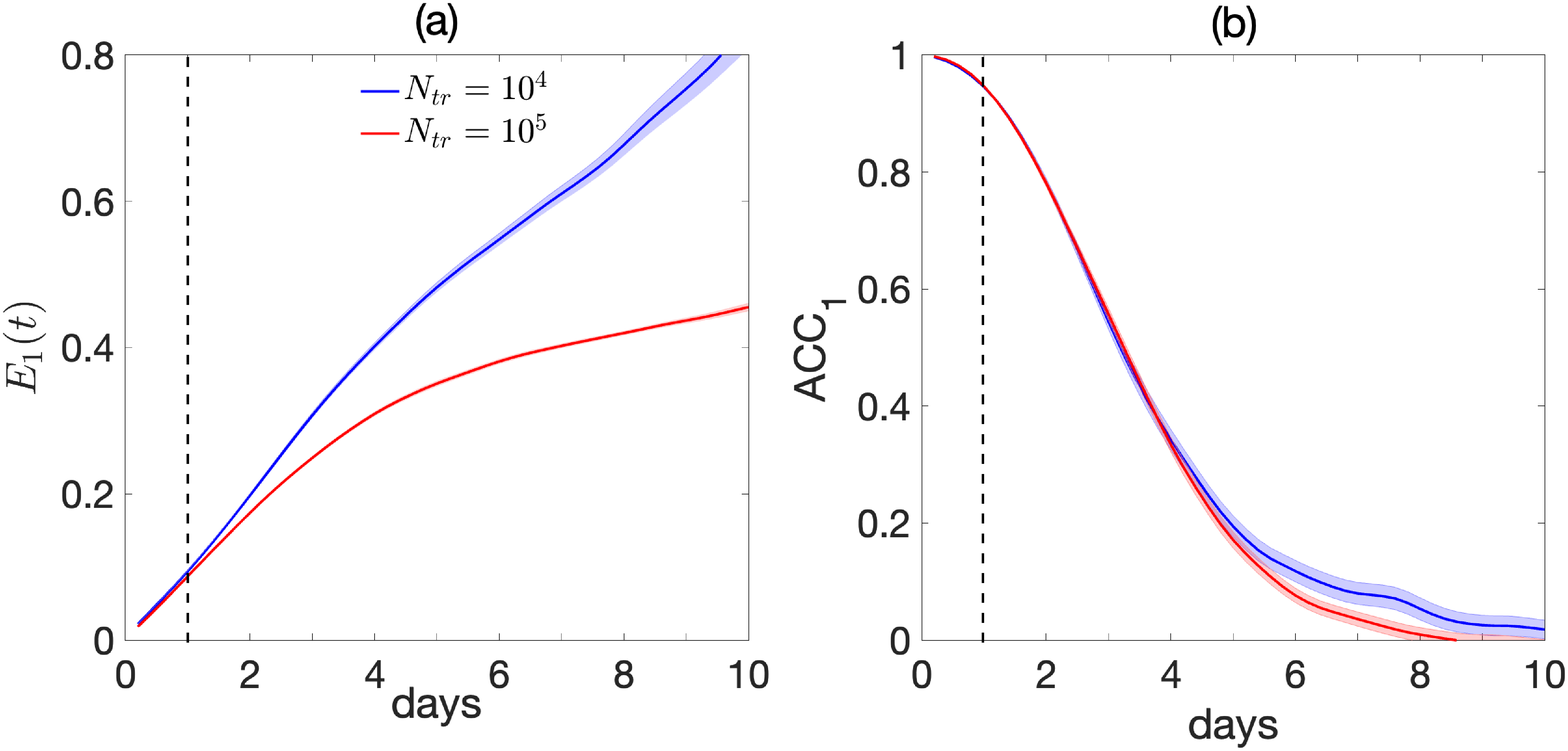}
\caption{\label{fig:DD_RMSE_ACC} Performance metrics of the U-NET trained on $N_{tr}=10^5$ and $N_{tr}=10^4$ samples. (a) $E_1(t)$, where the subscript ``1" refers to the first layer. (b) $\text{ACC}_1$ is the ACC for the first layer. Note that similar results are found for $E_2(t)$ and ACC for 2nd layer, but are not shown for brevity. Shading shows standard deviation over $100$ random noise-free initial conditions.}
\end{sidewaysfigure}

\TT{We quantify the accuracy of the predicted $\psi_1$ using the metrics defined in section~\ref{sec:metrics}.} Figure~\ref{fig:DD_RMSE_ACC} shows that $E_1(t)$ is $22.2 \%$ lower for the U-NET trained with $N_{tr}=10^5$ samples as compared to the one trained with $N_{tr}=10^4$ samples at day $3$. $\text{ACC}_1$ in Fig.~\ref{fig:DD_RMSE_ACC} shows that both U-NETs have roughly the same prediction horizon (time at which $\text{ACC}_1$ $\approx 0.60$).  

\TT{For the H-EnKF framework, we would need the U-NET to predict the states of the system for only $1$ day. For $1$-day prediction, both the relative error and ACC metrics of the U-NETs are not sensitive to $N_{tr}$. We have used an U-NET trained on $N_{tr} = 10^5$ samples as our data-driven model in the H-EnKF algorithm. However, using an U-NET trained on $N_{tr}=10^4$ samples, we would obtain the same performance for the H-EnKF algorithm.} We have also conducted experiments where an energy-constrained loss function was used to train the U-NET and found no significant improvement over the baseline performance in short-term (not shown for brevity).

\subsection{Performance of H-EnKF for DA}
\label{sec:HENKF_perf}

Next, we show the performance of H-EnKF as compared to standard EnKF for different values of $n_D$ and $n_N$. For $M_D$ in H-EnKF, we have used a U-NET trained on $N_{tr}=10^5$ samples. For regular EnKF, we assume that the dynamical model is the numerical solver of the QG system. As discussed in section~\ref{sec:EnKF}, we have considered the observation noise, $\sigma_{obs}=0.1$, to be $10\%$ of the standard deviation ($\approx 1.0$) of $\psi_k(t)$. Note that the computational cost (based on the wall-clock time of execution, see section~\ref{sec:cost_analysis} for details) of evolving $1$ numerical ensemble member ($n_N=1$) of the state vectors $\psi_k(t)$ for one $\Delta t$ is similar to evolving $200$ data-driven ensemble members ($n_D=200$) using the U-NET.
\begin{sidewaysfigure}
\includegraphics[width=\textwidth,angle=0,trim={0cm 0.0cm 0cm 0cm},clip]{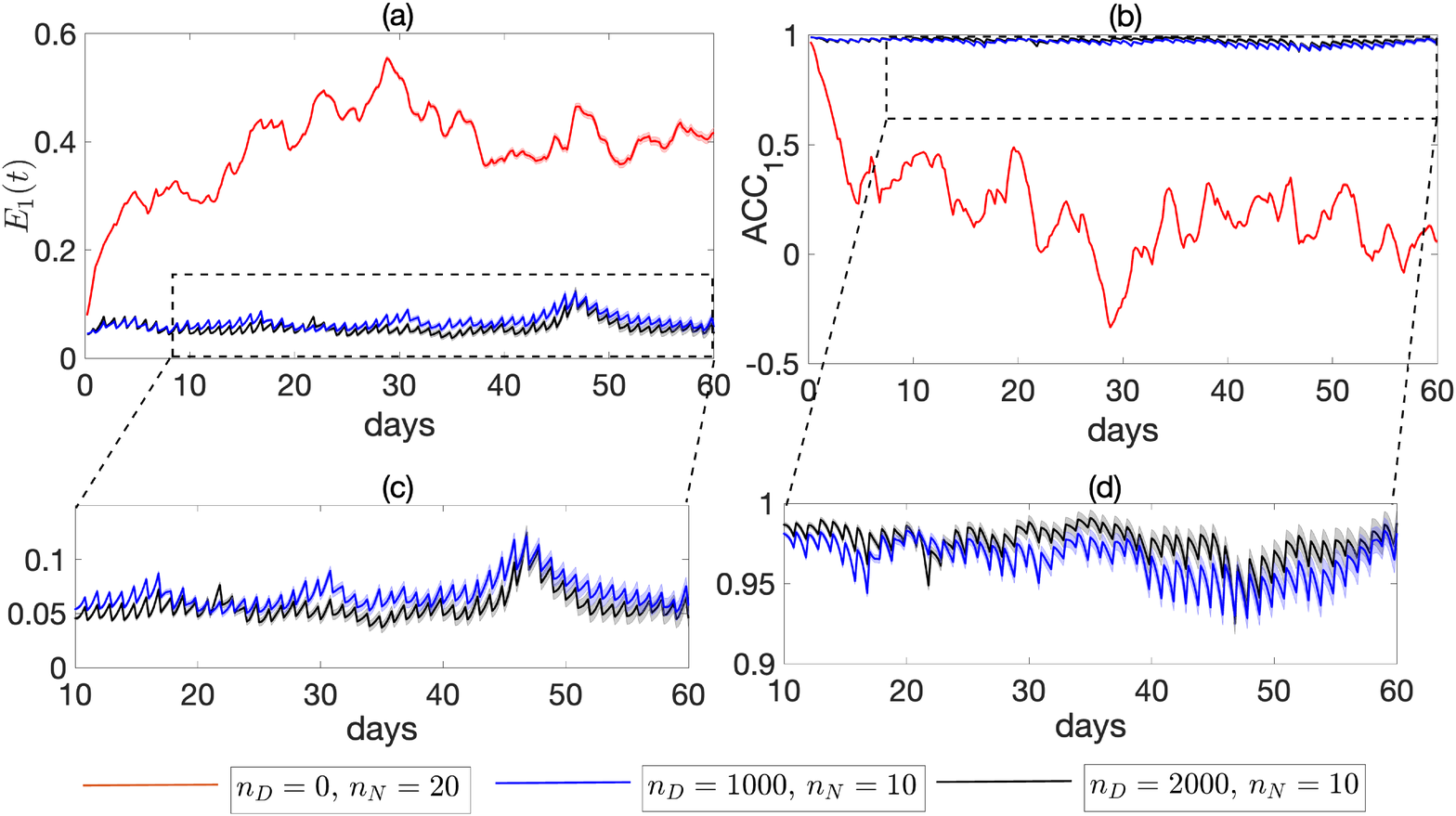}
\caption{Performance of H-EnKF and standard EnKF for $\psi_1(t)$ over $60$ DA cycles. (a) $E_1(t)$ over $60$ DA cycles for H-EnKF ($n_D=2000$, $n_N=10$), H-EnKF ($n_D=1000$, $n_N=10$), and EnKF ($n_D=0$, $n_N=20$). (b) Same as (a) but for $\text{ACC}_1$. (c) Zoomed in view of (a) betweeen $0\leq E_1(t) \leq 0.15$. (d) Zoomed in view of (b) between $0.90 \leq \text{ACC}_1 \leq 1.0$. Shading shows standard deviation over $30$ random initial conditions. $\sigma_b=0.80$ has been used for EnKF while $\sigma_b=0.10$ has been used for the H-EnKF models. The $\sigma_b$ value has been chosen to minimize the average $E_1(t)$ over $60$ DA cycles based on extensive trial and error.  }
\label{fig:H-ENKF_perf}
\end{sidewaysfigure}

Figure~\ref{fig:H-ENKF_perf}(a) and (c) show that the best performance is given by H-EnKF ($n_D=2000$, $n_N=10$) followed by H-EnKF ($n_D=1000$, $n_N=10$) and finally EnKF ($n_D=0$, $n_N=20$). Regular EnKF diverges due to the small number of ensemble members, a well known problem with EnKF in the absence of localization. Based on the computational cost \TT{analysis,} EnKF ($n_D=0$, $n_N=20$) is as expensive as H-EnKF ($n_D=2000$, $n_N=10$) while \TT{the latter} has $5\times$ smaller average $E_1(t)$ over $60$ DA cycles. Moreover, H-EnKF ($n_D=1000$, $n_N=10$) is $0.75\times$ cheaper than EnKF ($n_D=0$, $n_N=20$)  with $3 \times$ smaller average $E_1(t)$ over $60$ DA cycles.  

A similar conclusion can be made from Fig.~\ref{fig:H-ENKF_perf}(b) and (d) where $\text{ACC}_1$ for H-EnKF algorithms remain $\approx 0.95$ throughout the $60$ DA cycles while EnKF ($n_D=0$, $n_N=20$) shows a \TT{rapid decrease in $\text{ACC}_1$ from the beginning of the DA cycles \TT{(when localization is not performed)}. For H-EnKF, $\sigma_b=0.10$ has been used to obtain the best performance. We have conducted several trials with different values of $\sigma_b$ and chose the value that minimized the average $E_1(t)$ over $60$ DA cycles.}

\TT{These results demonstrate the effectiveness of the H-EnKF framework in terms of estimating a better initial condition from noisy observations of the system. In this framework, one can trade off a small number of computationally expensive numerical ensemble members with a large number of cheap data-driven ensemble members to improve the accuracy of the estimated analysis states without affecting the overall computational cost.}

It must be kept in mind that the performance of standard EnKF can be improved by simply increasing the number of numerical ensemble members or with localization. In our experiments with QG, we have seen that stable and divergence-free filters can be obtained with $O(1000)$ numerical ensemble members. \TT{Figure~\ref{sup_fig: EnKF} in~\ref{sec:enkf_large} shows the performance of EnKF over $10$ DA cycles with $n_N=1000$ and $n_N=5000$. For standard EnKF, $\sigma_b=0.80$ has been used to obtain the best performance. Similar to H-EnKF, we had conducted several trials to obtain the best $\sigma_b$, that minimized the average $E_1(t)$ over $10$ DA cycles. However, evolving $O(1000)$ ensemble members with the numerical solver comes at $50\times$ more computational cost compared to H-EnKF ($n_D=2000$, $n_N=10$). Moreover, for other practical systems with higher dimensions, evolving such a large size of ensembles may not even be computationally tractable. We have further demonstrated results with EnKF ($n_D=0, n_N=20$) with localization in Fig.~\ref{sup_fig:EnKF_with_local} (\ref{append:localization}), which shows that localization can also improve the performance of EnKF in this system without any increase in the computational cost. However, in more complex systems, localization can remove long-range physical correlations and affect the accuracy of the estimated analysis states \citep{miyoshi201410}. Furthermore, for other types of ensemble-based DA algorithms, such as particle filters, one may not be able to perform localization \citep{fearnhead2018particle,van2019particle}. Moreover, in particle filters, the large data-driven ensembles would provide an advantage in terms of sampling non-parametric and non-Gaussian distributions.}

\subsection{Analysis of Covariance Matrices}

\begin{figure*}[ht]
\includegraphics[width=\textwidth,angle=0,trim={4cm 10.0cm 25cm 0cm},clip]{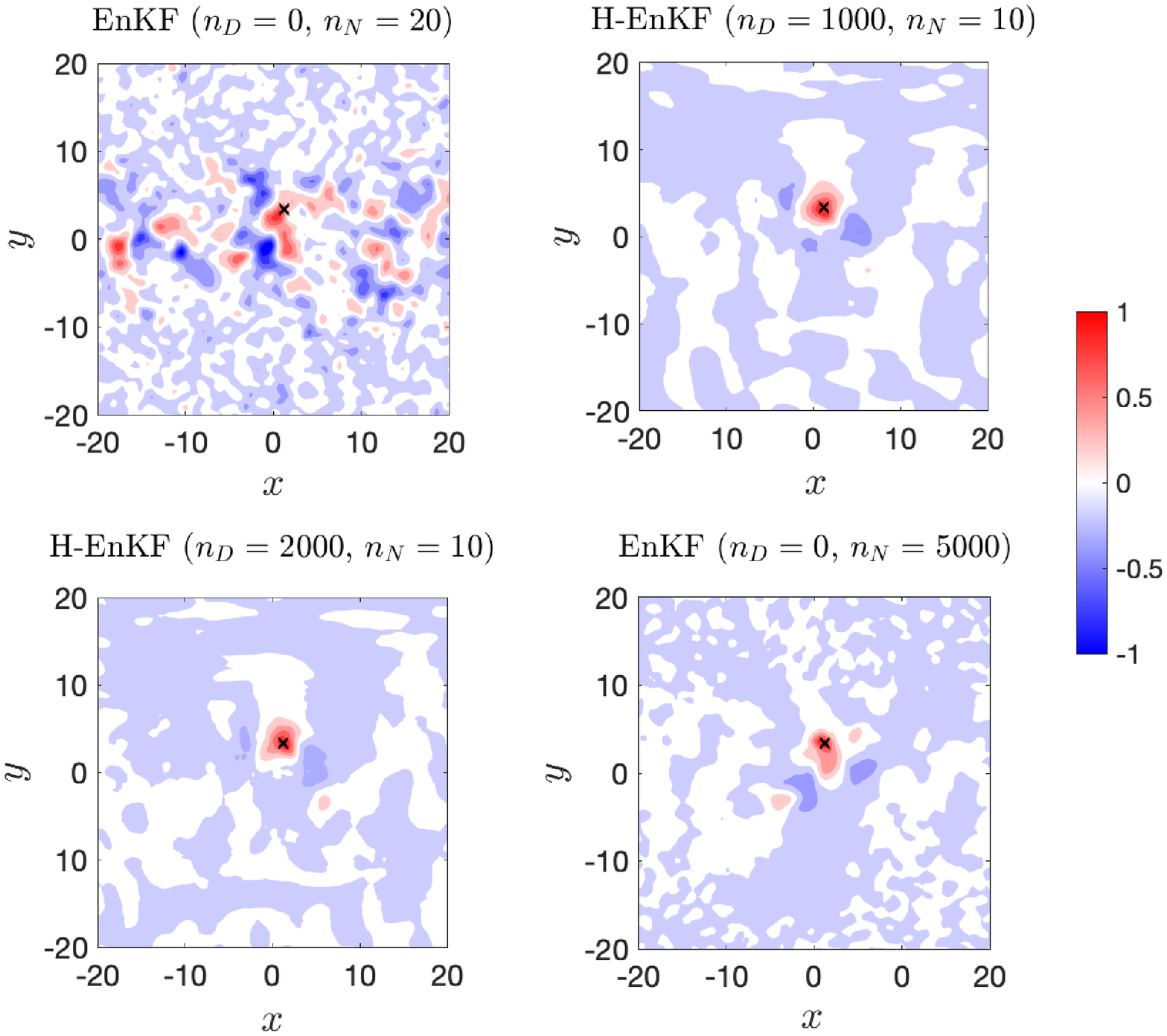}
\caption{The normalized background covariance matrix averaged over $60$ DA cycles, except EnKF ($n_D=0$, $n_N=5000$) which is averaged over $10$ DA cycles (due to limitations on computational cost). H-EnKF algorithms show more localized covariance structure and less spurious long-range correlation as compared to EnKF ($n_D=0$, $n_N=20$). The non-local structure of covariance of EnKF is due to low ensemble size leading to sampling error.}
\label{fig:covariance}
\end{figure*}

The \TT{superior} performance of H-EnKF ($n_D=1000$, $n_N=10$) and H-EnKF ($n_D=2000$, $n_N=10$) over EnKF ($n_D=0$, $n_N=20$) is due to the improved representation of the background error covariance matrix, $\mathbf{P_D}$, which is calculated using the $n_D$ data-driven ensembles. Figure~\ref{fig:covariance} shows that the $60$ DA cycles-averaged covariance structure of grid point $x=3.38$ and $y=1.21$ is localized around \TT{that} grid point for H-EnKF ($n_D=1000$, $n_N=10$) and H-EnKF ($n_D=2000$, $n_N=10$). However, for EnKF ($n_D=0$, $n_N=20$), the covariance structure is non-local with spurious long-range correlations across the domain. This is due to the reduction of sampling error during the computation of $\mathbf{P_D}$ in Eq.~(\ref{eq:P_b_HENKF}) using the $n_D$ ensembles. We have also shown the $10$ DA cycles-averaged covariance structure of EnKF ($n_D=0$, $n_N=5000$). Due to high computational cost, we have only performed 10 DA cycles with $n_N=5000$ ensembles. For this problem ($\approx 36000$ states), one can consider this covariance matrix to be close to the true covariance. It is clear that the H-EnKF covariance matrices are similar to the true covariance but at significantly lower computational cost.

\subsection{Performance of Free Prediction}
\begin{sidewaysfigure}

\includegraphics[width=\textwidth,angle=0,trim={1cm 0.0cm 2.2cm 0cm},clip]{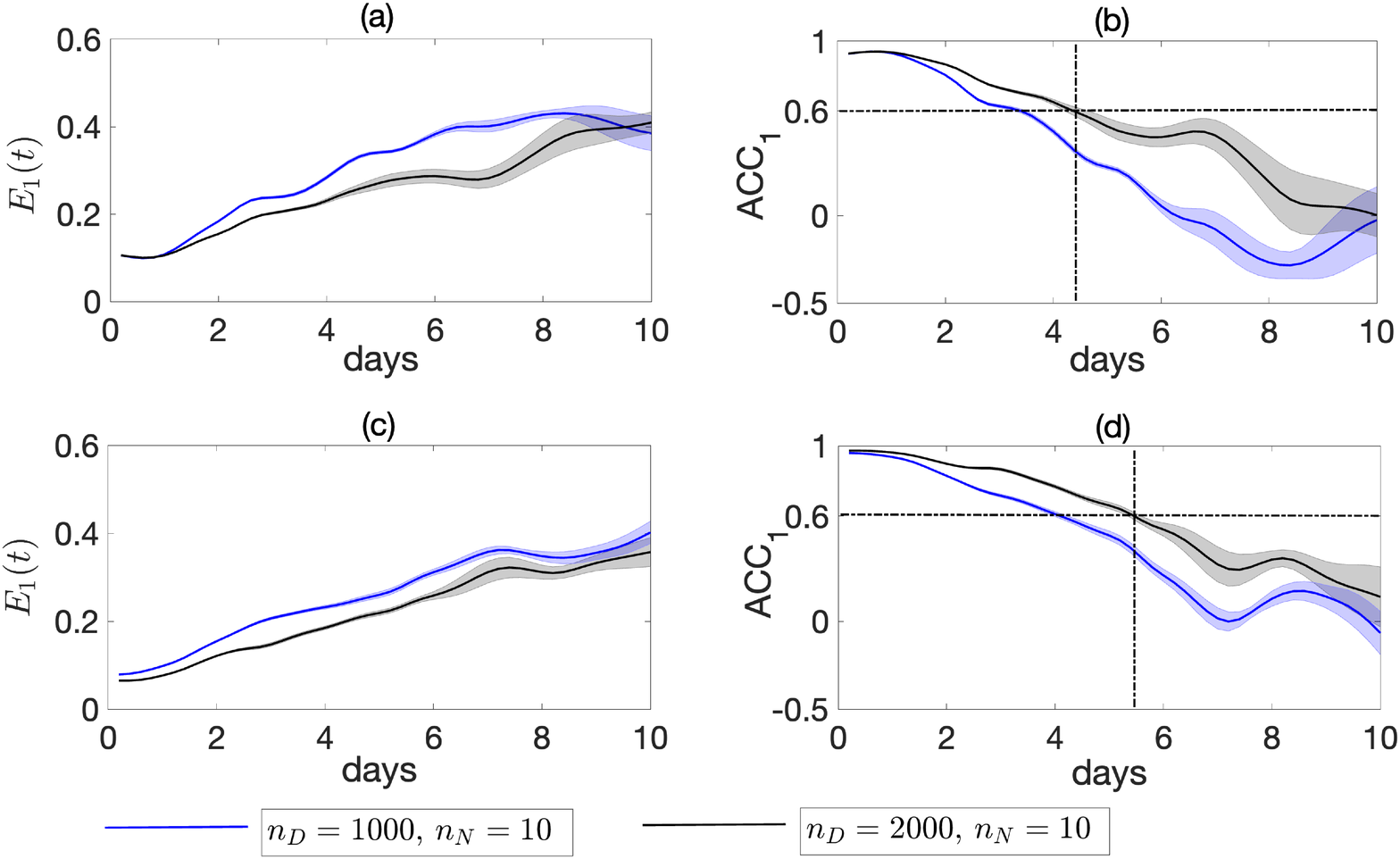}

\caption{Performance of free prediction of the QG numerical solver with initial condition (as mean analysis state) from H-EnKF. (a) $E_1(t)$ with initial condition at the end of $60$ DA cycles. (b) $\text{ACC}_1$ with the same initial condition as (a). (c) $E_1(t)$ with initial condition at the end of $10$ DA cycles. (d) $\text{ACC}_1$ with the same initial condition as (c). Shading shows standard deviation over $30$ random initial conditions. Performance of initial condition from EnKF ($n_D=0$, $n_N=20$) is not shown because it has no prediction skill at all.}
\label{fig:free_pred}
\end{sidewaysfigure}

Next, we compare the free prediction performance of the numerical model for the QG system with initial conditions that are obtained as the mean of the analysis states after 60 DA cycles (see Fig.~\ref{fig:H-ENKF_perf}). \TT{In Fig.~\ref{fig:free_pred}(a) and (b), we demonstrate that the initial condition obtained from H-EnKF ($n_D=1000$, $n_N=10$) has prediction skill up to $3.8$ days, while H-EnKF ($n_D=2000$, $n_N=10$) shows prediction skill up to $4.5$ days.} 

Figure~\ref{fig:free_pred}(c) and (d) show the free prediction of the numerical model for QG with initial conditions that are obtained as the mean of the analysis states after $10$ DA cycles. Here, the initial condition obtained from H-EnKF ($n_D=1000$, $n_N=10$) has prediction skill up to $4.1$ days and  H-EnKF ($n_D=2000$, $n_N=10$) has prediction skill up to $5.8$ days. 

\TT{We have not shown the free prediction performance of the initial condition from the standard EnKF because the analysis states obtained from the algorithm have too large of an error, as can be seen in Fig.~\ref{fig:H-ENKF_perf}, leading to no prediction skill at all.}

An important point to notice in Fig.~\ref{fig:free_pred} is the difference between the prediction skill of the initial condition obtained from H-EnKF ($n_D=1000$, $n_N=10$) and H-EnKF ($n_D=2000$, $n_N=10$). Although the initial conditions from both the models qualitatively seem to be very close, as is evident from inspecting $E_1(t)$ in Fig.~\ref{fig:H-ENKF_perf}, they result in significant difference in prediction skill ($\approx 0.7$ days for the initial condition at $60$ DA cycles and $1.7$ days for the initial condition at $10$ DA cycles). \TT{This shows that increasing $n_D$ (which is $200 \times$ cheaper than $n_N$) to compute $\mathbf{P_D}$ leads to significant improvement in the quality of the analysis states which in turn leads to an improvement in the prediction skill of the numerical model.} 

\subsection{Computational Cost Analysis and Scaling}
\label{sec:cost_analysis}

In this section, we discuss how the mean error over $10$ DA cycles, $\left<E_1(t)\right>$, where $\left<.\right>$ denotes mean over $10$ DA cycles, scales with the computational cost associated with different EnKF and H-EnKF algorithms. \TT{To have a fair cost comparison, we have executed both the EnKF and H-EnKF algorithms on the same hardware. The computations of background error covariance matrix, Kalman gain, and analysis states have been performed on an AMD EPYC 7742 CPU with $64$ cores for both EnKF and H-EnKF. The numerical model has been executed on the same CPU. However, the U-NET is run on a NVIDIA Tesla V$100$ GPU. We have not neglected the overhead cost of transferring data from GPU to CPU for H-EnKF. There are two things to note about the performance of the numerical simulation. First, one could have used a GPU-enabled numerical model as well, which would significantly improve its runtime performance. However, the overarching goal of this work is to facilitate DA in practical large-scale problems. In many practical problems, state-of-the-art simulation codes for fluid dynamics, combustion, weather prediction, etc., are CPU-based and would require enormous amounts of resources to refactor on GPUs. Second, one could have used a distributed-parallel numerical solver, which would also improve its runtime performance. However, one can also execute the U-NET on distributed GPUs as well, wherein the runtime performance gain due to distributed parallelism would be equivalent between the U-NET and the numerical solver. Owing to these careful observations and considerations, our computational cost analysis is fair.}  

Here, we define $1$ computational cost unit as the ``wall-clock" runtime of evolving $n_N=1$ ensemble member with the dynamical model over one $\Delta t$. Experimentally, we find that the runtime of $n_N=1$ ensemble member is the same as the runtime of $n_D=200$ ensemble members with the U-NET model over one $\Delta t$. To obtain a robust measure of runtime for the U-NET for one $\Delta t$, we had run the U-NET for $200 \Delta t$ and recorded the wall-clock time averaged over $100$ independent runs as $T^{D}_{200 \Delta t}$. From here, the runtime for the U-NET over $1 \Delta t$ is obtained as $T^{D}_{200 \Delta t}/200$. For the numerical solver, we had followed the same procedure to obtain the runtime for evolving one numerical ensemble member over one $\Delta t$ as $T^{N}_{200 \Delta t}/200$.

\begin{sidewaysfigure}
\includegraphics[width=\textwidth,angle=0,trim={0cm 0.0cm 0cm 0cm},clip]{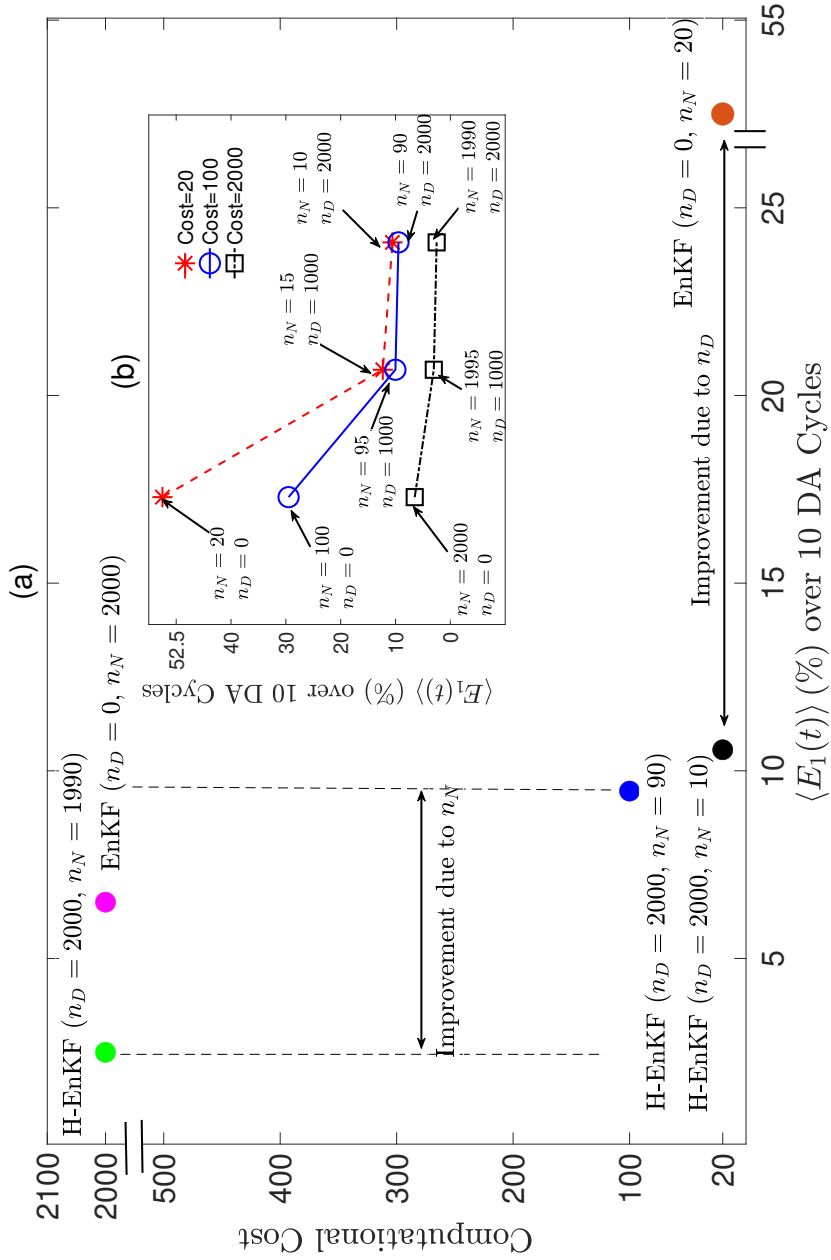}
\caption{Computational cost and accuracy scaling for H-EnKF and EnKF algorithms. Here, we have shown $\left<E_1(t)\right>$, the mean of $E_1(t)$, over $10$ DA cycles (instead of $60$) because of the high computational cost associated with computing $60$ DA cycles on EnKF models with large numerical ensembles ($O(1000)$). (a) $\left<E_1(t)\right>$ of different H-EnKF and EnKF algorithms along with their computational cost. (b) Scaling of $\left<E_1(t)\right>$ at different levels of fixed computational cost for different H-EnKF and EnKF models. The unit for computational cost is the runtime for the numerical QG solver to evolve one numerical ensemble for one $\Delta t$, $T^{N}_{200 \Delta t}/200$. See section~\ref{sec:cost_analysis} for more details.}
\label{fig:comp_cost}
\end{sidewaysfigure}

In Fig.~\ref{fig:comp_cost}(a), EnKF ($n_D=0$, $n_N=20$) shown with red circle has $\left<E_1(t)\right>$ of $52.6\%$ with computational cost of $20$ units. H-EnKF ($n_D=2000$, $n_N=10$) shown with black circle has a factor of $5$ smaller error at the same computational cost. By increasing the number of numerical ensemble members to $n_N=90$ and keeping the same number of data-driven ensemble members ($n_D=2000$), $\left<E_1(t)\right>$ goes down by a factor of $1.07$ but at $5$ times higher computational cost. With a further increase in cost to $2000$ units, $\left<E_1(t)\right>$ of EnKF ($n_D=0$, $n_N=2000$) improves by a factor $1.5$ in comparison to H-EnKF ($n_D=2000$, $n_N=10$). At the same cost, $\left<E_1(t)\right>$ of H-EnKF ($n_D=2000$, $n_N=1990$) improves by a factor of $4.0$. \TT{From this analysis, it is quite clear that trading a small number of numerical ensemble members for a large number of data-driven ones leads to a significant improvement in performance without an increase in computational cost.}

Figure~\ref{fig:comp_cost}(b) shows the scaling of $\left<E_1(t)\right>$ for H-EnKF and EnKF having different levels of fixed computational cost. We see that the effect of the background forecast derived from $n_N$ is more pronounced when $n_N$ is large ($n_N=2000$) (given by the black squares). Keeping the cost fixed at $2000$ units, adding more $n_D$ ($n_D=0$ to $n_D=2000$), decreases the error only very slightly (from $6.5\%$ to $2.5\%$). For low $n_N$ ($n_N=100$ or $n_N=20$), adding more $n_D$ significantly improves the performance of H-EnKF, shown by the blue circles (from $29.5\%$ to $9.46\%$) or red asterisk (from $48\%$ to $9.40\%$). It must be noted that H-EnKF ($n_D=2000$, $n_N=10$) at $5\times$ smaller cost performs almost equally well as H-EnKF ($n_D=2000$, $n_N=90$). A further improvement in the error (about $4\times$) is seen only with a $100 \times$ increase in computational cost. \\

\section{Discussion and Summary}
\label{sec:discussion}
In this paper, we have proposed H-EnKF: a hybrid ensemble Kalman filter algorithm that leverages a deep learning-based data-driven model to efficiently evolve a large ensemble of states of a dynamical model to better estimate the background error covariance with low sampling error. The ensemble of background forecast states used in the H-EnKF algorithm is obtained from an accurate, high-resolution, numerical solver evolving a small number of ensemble of states (owing to computational cost). This combination allows one to obtain an accurate DA algorithm at low computational cost alleviating one of the major sources of error --sampling error in covariance due to low ensemble size-- without the need for localization. 

\TT{There is a major difference between how the numerical ensembles are evolved throughout the DA cycles as compared to the data-driven ensembles in H-EnKF. In H-EnKF, after every DA cycle, the data-driven ensembles are regenerated with Gaussian noise added to the mean of the analysis states. This is due to a limitation of current data-driven models for high-dimensional systems, which become unstable when evolved for a long period of time \citep{chattopadhyay2020deep,chattopadhyay2022long,keisler2022forecasting}. We acknowledge that in more complex systems, this approach can cause issues, as frequently perturbing the state can violate conservation laws, e.g., of mass (this is not a problem here as we are perturbing streamfunctions, $\psi$). In such problems, the perturbation should be added such that they do not violate the conservation laws, e.g., see Zeng \textit{et al.}~\cite{zeng2016study}}.

We have shown that the H-EnKF algorithm achieves a stable filter without localization and has a factor of $5$ smaller error at the same computational cost as that of standard EnKF for the two-layer QG system (Fig.~\ref{fig:H-ENKF_perf}). We have further shown that H-EnKF is most useful for situations where one can only afford small numerical ensembles (Fig.~\ref{fig:comp_cost}). In such situations, one can trade off a small number of numerical ensemble members for a larger number of data-driven ones and obtain a more accurate covariance estimation. \TT{In H-EnKF applied to the two-layer QG system, the runtime for the evolution of the data-driven ensembles is a factor $200$ smaller than the numerical ones. However, in more practical problems, such as weather prediction, the difference in runtime between evolving data-driven and numerical ensembles  can be much larger, e.g., in FourCastNet \cite{jpathak2022}, the data-driven model is $100000 \times$ faster than the state-of-the-art operational weather forecasting models. We expect the H-EnKF algorithm to be most effective for high-dimensional problems such as these.}

As a test case, the proposed hybrid framework has been applied to EnKF, a specific ensemble-based DA algorithm that is used with Gaussian observation noise. However, one can readily extend this hybrid framework to other types of novel extensions of regular EnKF, e.g., multi-model EnKF \citep{bach2022multi} or other ensemble-based DA algorithms, e.g., particle filters. Particle filters are especially useful for systems with strong non-Gaussian observation noise. However, due to large computational cost, it has been difficult to use in high-dimensional systems such as geophysical flows \cite{wan2001unscented,fearnhead2018particle}. \TT{For future work, we aim to explore how the H-EnKF algorithm can be used in particle filter-based DA.} 

The EnKF algorithms with small ensemble members are currently used in weather prediction by leveraging ad-hoc techniques such as localization that removes spurious long-range correlations. Localization can also remove physical correlations in the flow fields as well, which would result in inaccuracies in the covariance \cite{miyoshi201410}. Due to low computational cost during inference, deep learning-based surrogates can generate a large ensemble of states that can better approximate the true background covariance as compared to artificially obtained covariance with localization. \TT{However, for applications where localization works well, one can use localization on the covariance matrix obtained from the data-driven ensembles in our hybrid framework. In such scenarios, we can bring down the computational cost even further by evolving only a small number of data-driven ensembles (which are already quite inexpensive to evolve). Moreover, localization may be difficult to perform on other ensemble-based DA algorithms, e.g., particle filters. For such algorithms, our hybrid approach which does not require localization would be useful as well.} 

One of the key aspects of the H-EnKF framework is the accurate data-driven surrogate which needs to be built for every application. However, building a fully data-driven surrogate for high-dimensional realistic systems, e.g., weather and climate,  can be challenging. Recent successes in data-driven short-term weather modeling \citep{weyn2020improving,chattopadhyay2021towards,jpathak2022} show that short-term accurate surrogates can be as accurate as operational numerical models if trained on observations \citep{jpathak2022}. Further improvement in data-driven surrogates can be achieved by conserving key physics as can be seen in many different applications in both fluids \citep{guan2022learning} and weather and climate community \citep{kashinath2021physics}.

\section{Acknowledgements}

We thank Matti Morzfeld and Yonquiang Sun for insightful comments and discussions. This work was supported by an award from the ONR Young Investigator Program (N00014-20-1-2722) and a grant from the NSF CSSI program (OAC-2005123) to P.H. Computational resources were provided by NSF XSEDE (allocation ATM170020) to use Bridges GPU and the Rice University Center for Research Computing. The codes for H-EnKF are publicly available at \url{https://github.com/ashesh6810/Hybrid-Ensemble-Kalman-Filter}.

\appendix

\section{Equations for Error metrics}
\label{secA:error}
We define relative error as:

\begin{eqnarray}
\label{eq:RMSE}
E_k(t)  = \frac{|| \psi_k^{pred}(t)-\psi_k^{true}(t) ||_2}{max\left(\psi_k^{true}(t)\right)}.
\end{eqnarray}
$\psi_k^{pred}$ is the predicted streamfunction and $\psi_k^{true}$ is the true streamfunction obtained from numerical simulations.

We define $\text{ACC}_k$ as:
\begin{eqnarray}
\label{eq:ACC}
\text{ACC}_k = \frac{\mathlarger{\mathlarger{\Sigma_m  \Sigma_p}}\left(\left(\psi_{k,m,p}^{pred}(t)-\left<\psi_{k,m,p}^{true}\right>\right)\times \left(\psi_{k,m,p}^{true}(t)-\left<\psi_{k,m,p}^{true}\right>\right)\right)}{\sqrt{\left(\mathlarger{\mathlarger{\Sigma}} _m\mathlarger{\mathlarger{\Sigma}}_p\left(\psi_{k,m,p}^{pred}(t)-\left<\psi_{k,m,p}^{true}\right> \right)^2\times\mathlarger{\mathlarger{\Sigma}} _m\mathlarger{\mathlarger{\Sigma}}_p\left(\psi_{k,m,p}^{true}(t)-\left<\psi_{k,m,p}^{true}\right> \right)^2\right)}},
\end{eqnarray}
where $\left<\psi_{k,m,p}^{true}\right>$ is the time-averaged value of $\psi_{k,m,p}^{true} (t)$ and the indices $m$ and $p$ refer to the latitudinal and longitudinal grid points on which $\psi_k(t)$ is represented. Similar to $E_k(t)$, ACC is also computed for each layer.


\section{EnKF with Large Numerical Ensembles}
\label{sec:enkf_large}
\begin{sidewaysfigure}

\includegraphics[width=\textwidth]{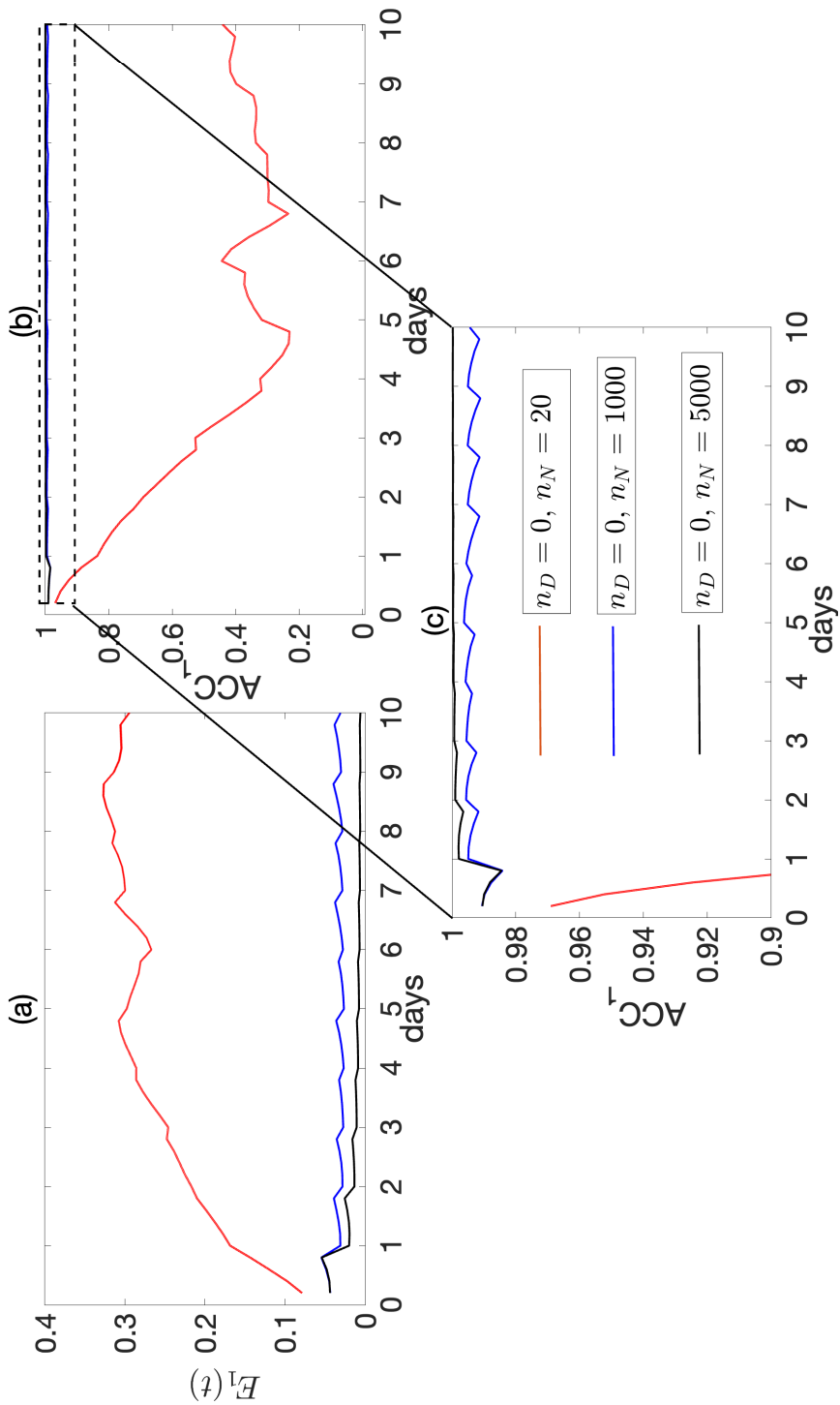}
\caption{Performance of EnKF for $\psi_1(t)$ over $10$ DA cycles with large $n_N$. Due to the computational cost of evolving large ensembles, we only report $E_1(t)$ and $\text{ACC}_1$. (a) $E_1(t)$ for EnKF ($n_D=0$, $n_N=20$), EnKF ($n_D=0$, $n_N=1000$), and EnKF ($n_D=0$, $n_N=5000$). (b) Same as (a) but for the $\text{ACC}_1$. (c) Zoomed in view of (b) between $0.90 \leq \text{ACC}_1 \leq 1.0$. Shading shows standard deviation over $30$ initial conditions. $\sigma_b=0.80$ has been used for all the EnKF algorithms. The $\sigma_b$ value has been chosen to minimize the average $E_1(t)$ over $10$ DA cycles based on extensive trial and error.}
\label{sup_fig: EnKF}
\end{sidewaysfigure}

Here, we show that EnKF with a large number of numerical ensemble members ($n_N$) can have stable and divergence-free filters with low $E_1(t)$. However, in order to have performance comparable to H-EnKF, one needs to have $O(1000)$ $n_N$ which comes at a high computational cost. Figure~\ref{sup_fig: EnKF} shows that as $n_N$ increases, the performance of EnKF improves as well. This is expected since with large ensembles, the matrix, $\mathbf{P}$, would not suffer from sampling error and spurious correlations, as shown in Fig.~\ref{fig:covariance} as well.  

\section{EnKF with Small Numerical Ensembles and Localization}
\label{append:localization}
\begin{sidewaysfigure}

\includegraphics[width=\textwidth,angle=0,trim={0cm 6.0cm 0cm 0cm},clip]{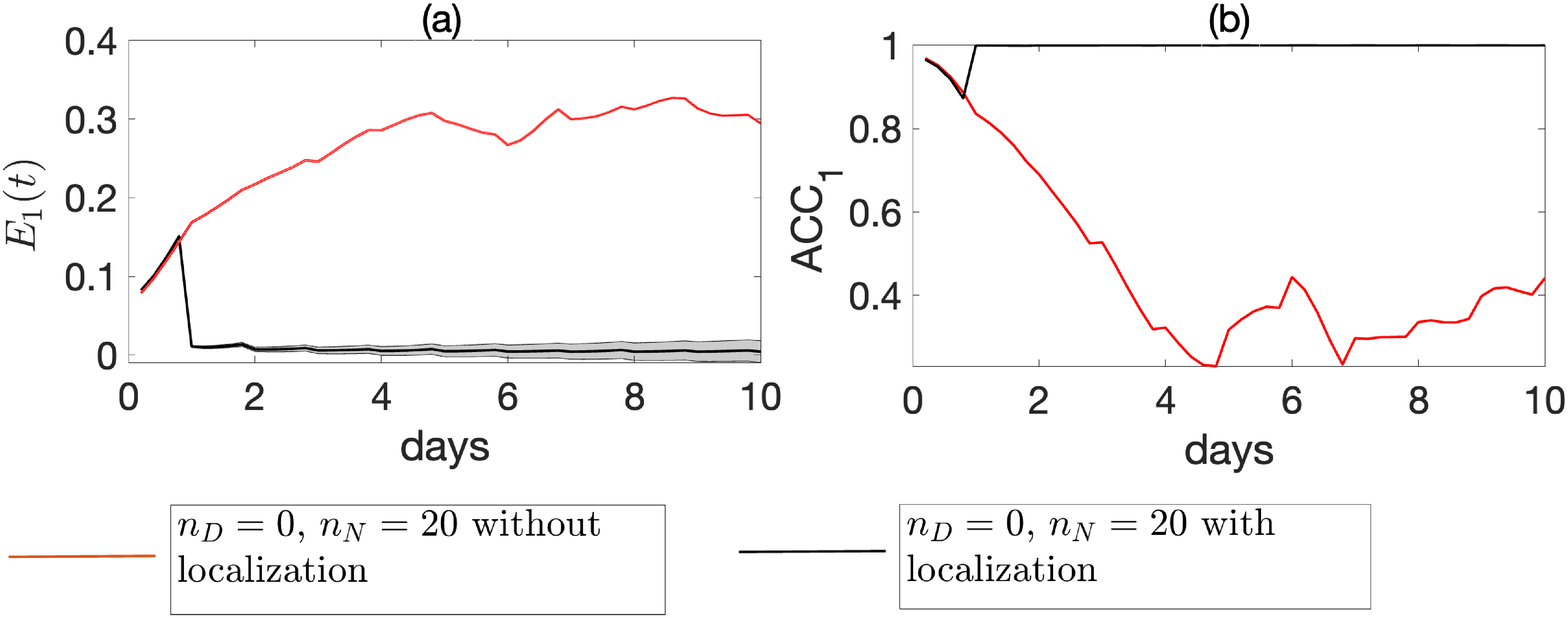}
\caption{Performance of EnKF for $\psi_1(t)$ over $10$ DA cycles with localization. (a) $E_1(t)$ over $10$ DA cycles for EnKF ($n_D=0$, $n_N=20$) and EnKF ($n_D=0$, $n_N=20$) with localization. (b) Same as (a) but for $\text{ACC}_1$. Shading shows standard deviation over $30$ initial conditions. $\sigma_b=0.80$ has been used for both the EnKF algorithms. The $\sigma_b$ value has been chosen to minimize the average $E_1(t)$ over $10$ DA cycles based on extensive trial and error.}
\label{sup_fig:EnKF_with_local}
\end{sidewaysfigure}

In Fig.~\ref{sup_fig:EnKF_with_local}, we report $E_1(t)$ and $\text{ACC}_1$ for EnKF ($n_D=0, n_N=20$) with and without localization. \TT{Here, we have used covariance localization~\citep{asch2016data}. The Gaspari-Cohn function is used to generate the regularizing correlation function. A radius of $5l$, where $l$ is the $L_2$ distance between two consecutive grid points, has been used as the radius in the Gaspari-Cohn function.} In this system, localization is an effective method to obtain divergence-free filters with similar performance as H-EnKF ($n_D=2000$, $n_N=20$).  However, it must be noted that localization often removes long-range physical correlations as well and is required to be tuned for each application in more complex systems \citep{miyoshi201410}.

\bibliographystyle{elsarticle-num}
\bibliography{main}

\end{document}